\def\year{2020}\relax
\documentclass[letterpaper]{article} 
\usepackage{aaai20}  
\usepackage{times}  
\usepackage{helvet} 
\usepackage{courier}  
\usepackage[hyphens]{url}  
\usepackage{graphicx} 
\usepackage{hhline}
\usepackage{graphicx}  
\usepackage{longtable}

\title{Predicting Illness for a Sustainable Dairy Agriculture: \\Predicting and Explaining the Onset of Mastitis in Dairy Cows}
\author{Cathal Ryan,\textsuperscript{\rm 1,2} Christophe Guéret\textsuperscript{\rm 3}, Donagh Berry,\textsuperscript{\rm 4,2} \\ \Large \textbf{Medb Corcoran\textsuperscript{\rm 3}, Mark T. Keane,\textsuperscript{\rm 1,2} Brian Mac Namee\textsuperscript{\rm 1,2}} \\ 
\textsuperscript{\rm 1}School of Computer Science, University College Dublin, Belfield, Dublin 4, Ireland \\ 
\textsuperscript{\rm 2}VistaMilk, Science Foundation Ireland Research Centre \\
\textsuperscript{\rm 3}Accenture Labs, Dublin \\
\textsuperscript{\rm 4}Teagasc, Animal \& Grassland Research and Innovation Centre, Moorepark, Fermoy P61 P302, Co. Cork, Ireland 
}
 
\begin{document}

\maketitle

\begin{abstract}
\begin{quote}

 Mastitis is a billion dollar health problem for the modern dairy industry, with implications for antibiotic resistance. The use of AI techniques to identify the early onset of this disease, thus has significant implications for the sustainability of this agricultural sector. Current approaches to treating mastitis involve  antibiotics and this practice is coming under ever increasing scrutiny. Using machine learning models to identify cows at risk of developing mastitis and applying targeted treatment regimes to only those animals promotes a more sustainable approach. Incorrect predictions from such models, however, can lead to monetary losses, unnecessary use of antibiotics, and even the premature death of animals, so it is important to generate compelling explanations for predictions to build trust with users and to better support their decision making. In this paper we demonstrate a system developed to predict mastitis infections in  cows and provide explanations of these predictions using counterfactuals. We demonstrate the system and describe the engagement with farmers undertaken to build it. 
\end{quote}
\end{abstract}

\section{Introduction}
If agriculture is to have a future it needs to be sustainable; for the dairy sector this requires zero-carbon farming (using pasture rather than indoor systems) and minimising the use of antibiotics (to avoid resistances arising).  Artificial intelligence (AI) has a major role to play in these efforts (see, for example, the UN AI for Good initiative), and our work has focused on a major animal health problem, mastitis, for dairy farms (which costs US\$19-32B per annum \cite{Eckersall2016}).  Current approaches to handling mastitis involve the administration of large quantities of antibiotics, which is not sustainable \cite{Eckersall2016}. The proliferation of sensors on modern dairy farms, and the data that they generate, offers an opportunity to more sustainably address the mastitis issue. Data collected from farms can be used to build machine learning models that proactively identify cows likely to develop mastitis \cite{anglart}
, enabling farmers to treat those cows in targeted ways. 

Predicting a cow's likelihood to develop mastitis, however, has significant consequences. If infections are missed then farmers will lose significant income through discarded milk \cite{halasa_2007} and this may even lead to increased chances of early culling of a cow \cite{hogeveen_huijps_lam_2011}. If healthy cows are flagged as mastitis risks, farmers are likely to forego milking income and administer antibiotics unnecessarily. Farmers can also use the output of a prediction model to gain insights about how to improve the health status of their herd by identifying contributors to mastitis infection. This opportunity, and the significant consequences of actions taken by farmers based on model output, mean it is very important that the output of prediction models used is explainable. 

In this paper we describe a machine learning model built to predict the propensity of a cow to develop mastitis, and an approach based on counterfactuals to generate explanations of the predictions made by the model. Moreover, we describe how engagement with stakeholders in the farming community guided the design of this system, particularly in identifying  features that could actually be impacted and appropriate levels of change for use in the counterfactual system. We demonstrate the effectiveness of this approach with worked examples and suggest directions for future work.


\section{Related Work}\label{sec:related_work}

Machine learning has been applied to the agricultural domain for tasks ranging from grass growth prediction \cite{10.1007/978-3-030-29249-2_12}, to grass phenotype identification \cite{Narayanan2020}, to disease detection within cows \cite{Mammadova2015ApplicationON}. In the latter case predicting mastitis has received a significant amount of attention from researchers with approaches based on simple statistical models (eg. \cite{Khatun2018DevelopmentOA}) and more sophisticated approaches using machine learning (e.g. \cite{Sun2010DetectionOM}). Approaches can be distinguished by the type of input data they use, with some using simple features describing cows and aggregate milking behaviour \cite{anglart} and others using much more precise information such as measurements of milk yield at the level of an individual quarter of a cow's udder \cite{kamphuis_mollenhorst_feelders_pietersma_hogeveen_2010}. In this problem domain, there are considerable costs to miss-prediction: potentially infected cows may need to be quarantined, their milk cannot be used and expensive antibiotic treatment may be required. So, model predictions need to be coupled with compelling explanations to allow farmers to make actionable decisions.


Recent reviews of explainable AI have argued for the use of counterfactuals to explain the causal importance of different features involved in a model’s predictions \cite{adadi_berrada_2018,miller_2019}. 
For example, if the model makes a prediction that ``\emph{Cow \#42 will remain healthy}'' and the vet/farmer asks ``\emph{why?}'', then the explanation module can provide a counterfactual explanation saying
\begin{quote}\emph{If cow \#42 had a somatic cell count of 150 and a protein percentage of 5\% she would be likely to succumb to mastitis}. \end{quote}
\noindent This information enables the end-user to assess whether they think the critical differences between the proposed prediction and the counterfactual explanation warrant acceptance of the model’s directions.

The growth in research interest in counterfactuals is largely driven by computational, psychological and legal reasons. Computationally, many methods generate counterfactuals by perturbing features of the original instance \cite{wachter_mittelstadt_russell_2017}, 
or nearest unlike neighbor \cite{keane_smyth_2020}; which can be seen as a follow on from the popular LIME methods \cite{guidotti_monreale_giannotti_pedreschi_ruggieri_turini_2019,Ribeiro2018AnchorsHM}.  Psychologically, it has been shown that people can readily and naturally understand the causal roles of features using counterfactuals \cite{miller_2019,Byrne2019CounterfactualsIE,Byrne2007PrcisOT}. Legally, it has been argued that counterfactuals are GDPR compliant meeting the requirements for explaining automated decision-making \cite{Grath2018InterpretableCA}.   

\begin{figure}[t]
\centering
\includegraphics[width = 1.0\linewidth]{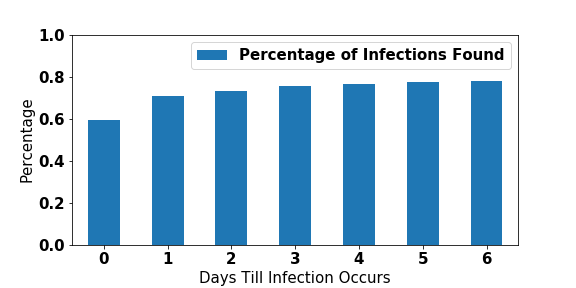}
\caption{Proportion of infections found by sub-clinical mastitis prediction model at different time horizons.}
\label{fig:counterfactuals}
\end{figure}

\section{Predicting Mastitis}\label{sec:predicting_mastitis}

We have developed a propensity model that uses the record of milking on a particular day, a history of recent milkings, and details of the cow herself to predict the likelihood that a cow will develop sub-clinical mastitis in the next 7 days. To build this model we use a dataset originating from 7 different research farms located around Ireland. This dataset contains details of over 2,300 cows over a period of 10 years and includes aspects of each cow's history including milk recordings, infection history, and genetic details. Milk recording data is the richest data source and includes the volume of milk yielded when a cow was milked; characteristic of that milk including fat, lactose and protein content; and somatic cell count (SCC) recorded from the collected milk, which is a useful indicator of mastitis \cite{sharma2011relationship}. Milk yield is available twice per day; fat, lactose, and protein content of milk and SCC are measured approximately every 7 days.


Based on the raw data available, we generate a set of features describing a cow on a particular day that form the basis of prediction. The milk yield and milk characteristic data form multi-variate time series describing a cow's history from which multiple features could be generated---for example minimum, maximum, mean, and skewness values of each variable over histories of 15 and 30 days. Any feature included in our prediction model, however, has the potential to be included in the counterfactual explanations generated to describe predictions. Including a model based on a large number of features derived from time series, however, could lead to contradictory and inactionable  counterfactual explanations, such as the following, being generated:
\begin{quote}
\textit{If Cow \#42 had a lower mean yield over the last 15 days and a higher minimum yield over the last 15 days she would be likely to succumb to mastitis}
\end{quote}


\noindent Therefore, in our model for each milk characteristic (\textit{e.g.} yield and lactose content), we use only the values from the current day (or the most recent values) plus a single historical aggregate: skewness over the  last 30 days. The features used in the model are shown in the first column of Table \ref{tab:features}. The model trained is a gradient boosting model and we use data from 2010--2018 to train it, and data from 2018--2019 to evaluate it. Figure \ref{fig:counterfactuals} illustrates the ability of the model to correctly predict sub-clinical mastitis cases at different time horizons from 1 day in advance to 7 days in advance (measured using the hold-out test set described above).  Domain experts informed us that capturing $70\%$ of infections at least 1 day in advance of their onset would lead to a model more useful than current approaches. The model trained achieves this from 5 days in advance.

\begin{table*}[!t]
\centering
{\renewcommand{\arraystretch}{1.1} 
\begin{tabular}[c]{@{}lllll@{}}
\hline
Feature Name & Actionable & Actionable Time & Confidence & Size of
Change\\
\hline
Somatic cell count (SCC) & No & & Very high
\\
Yield & Yes & & Low & 2 kg 
\\
Fat percentage & Yes & 1-2 days & Low & 0.05 percentage units
\\
Protein percentage & Yes & 1-2 days & Low & 0.05 percentage units
\\
Lactose percentage & No & & Low
\\
Urea & Yes & & Low
\\
Body condition score (BCS) & Yes & 2 weeks & Very high & 0.25 units
\\
Weight & Yes & 1 week & Medium & 10 kg 
\\
Genetic merit & Yes & 5 years & Very high & 
\\
Parity & No & & Very high
\\
Days since calving & No & & Very high
\\
Calendar month & No & & Low 
\\
Infections for this lactation stage (farm) & No & & Very high
\\
Infections for this lactation stage (cow) & No & & Very high
\\
Infections for this cow & No & & Very high
\\
\begin{tabular}[c]{@{}l@{}}Proportion of cows infected per farm \\ and calving year\end{tabular}& No & & Very high
\\
\hline
\end{tabular}}
\caption{The features used in the model and an indication of whether or not they are actionable and can be used to build confidence. For those that are actionable, the time within which changes can be made, and the level of change that is appropriate to suggest, also are provided.}
\label{tab:features}
\end{table*}



\section{Generating Useful Counterfactuals}\label{sec:generating_counterfactuals}
Given a trained binary classification model $f(\dots)$, counterfactuals represent a hypothetical input vector $\mathbf{x}'$ giving the output $f(\mathbf{x}') = y'$ where $y'$ is the \emph{positive} class. This vector $\mathbf{x}'$ is derived from an actual vector $\mathbf{x}$ as a set of small changes $\Delta{}\mathbf{x}$ to apply to $\mathbf{x}$ to change a \emph{negative} class outcome $f(\mathbf{x})=y$ into $y'$. That is, we seek to optimize $\Delta{}\mathbf{x}$ under the constraint $f(\mathbf{x} + \Delta{}\mathbf{x}) = y'$. In our implementation, $\Delta{}\mathbf{x}$ is found using the COBYLA~\cite{powell1994} constrained optimization approach. We use a Manhattan distance weighted feature-wise with the inverse median absolute deviation, as it promotes sparsity and is more robust to outliers.

For our use-case the \emph{positive} class is a prediction that a cow will succumb to sub-clinical mastitis so we seek to provide support for explaining why a given cow is predicted not to succumb to sub-clinical mastitis. Considering that users find explanations more useful when only a small number of changes appear \cite{keane_smyth_2020} we set a constraint to change no more than 3 features of the model. We further take a number of additional and domain-specific constraints into account as outlined in Table~\ref{tab:features}. These were established through consultation with a group of domain experts (farmers and animal health experts). For some of the features a minimum magnitude of change is imposed as, for instance, it is not meaningful to ask for milk yield to be changed by 0.1 litre per milking. All of the features come with an indication of \emph{actionability} and \emph{confidence}. The latter is, to the best of our knowledge, new to applied research on counterfactuals. Some features that are not actionable for the farmers could be still relevant to report on in the counterfactual in order to build confidence and trust in the model. The best example is somatic cell count (SCC): there is little the farmer can do about this metric in the short term but it is part of their own mental model for mastitis. Surfacing this feature as part of the explanation for the predicted outcome is expected to contribute to building a trust relationship between the farmer and the prediction system.

\section{Worked Example}\label{sec:worked_example}

We use the predictions made on our hold-out test set to demonstrate the ability of the system developed to predict mastitis and generate useful counterfactual explanations. We randomly sampled 10,000 test instances predicted to be healthy with a high prediction score (greater than 0.8) and generated counterfactual explanations for each of these predictions with the system described in the previous section. 

\begin{figure}[t]
\centering
\includegraphics[width = 0.65\linewidth]{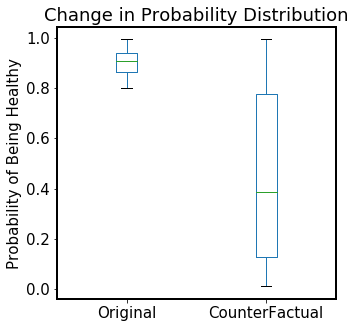}
\caption{Probability Changes}
\label{fig:Change}
\end{figure}

Figure \ref{fig:Change} summarises the prediction scores for the original model predictions and the prediction scores of  counterfactual instances generated. We can see that the new counterfactual instances consistently result in low prediction scores. Table \ref{tab:Change} shows a sample of  the counterfactuals generated (values in parentheses are those associated with the counterfactuals). The following text would be used to present the first counterfactual to a farmer:
\begin{quote}
\textit{If cow \#42 had an increase of one and a half units with respect to Yield she would be likely to succumb to mastitis.}
\end{quote} 
 We can see that only features marked as actionable or confidence building are changed, and that the changes are in the units specified as actionable. 

\begin{table}[t]
\centering
{\renewcommand{\arraystretch}{1.2} 
\begin{tabular}{@{} l r r @{}}
\hline
\begin{tabular}[c]{@{}l@{}}Feature Changed\end{tabular}       & $\Delta$ Feature & Prediction Score    \\ 
\hline
\begin{tabular}[c]{@{}l@{}} Yield \end{tabular} & 1.5 & 0.979 (0.368) \\ 
\begin{tabular}[c]{@{}l@{}} Fat\end{tabular} &  1 & 0.979 (0.192) \\  \hline
\end{tabular}}
\caption{Original versus counterfactual example}
\label{tab:Change}
\end{table}



\section{Conclusions \& Future Work}\label{sec:concluions}

In this paper we demonstrate a system built to predict mastitis in cows and generate explanations of its predictions using a counterfactual approach. We took input from stakeholders to understand which features could be actionable and which would help build confidence in the system's predictions. We demonstrate the ability of this system to create meaningful explanations while keeping the number of changes to the original data as small as possible. We are currently presenting this system to farmer focus groups for feedback. 

\section{ Acknowledgments}
This publication has emanated from research conducted with the financial support of Science Foundation Ireland (SFI) and the Department of Agriculture, Food and Marine on behalf of the Government of Ireland under Grant Number [16/RC/3835].
\bibliographystyle{aaai}
\bibliography{references}
\end{document}